\documentclass[conference]{IEEEtran}
\IEEEoverridecommandlockouts
\usepackage{tikz}
\usepackage{textcomp}
\usepackage{hyperref}
\usepackage{lipsum}

\newcommand\copyrighttext{%
  \footnotesize \textcopyright This work has been accepted as a conference paper at 2023 International Joint Conference on Neural Networks (IJCNN), Gold Coast, Australia, 2023, pp. 01-07, doi: 10.1109/IJCNN54540.2023.10191657.}
\newcommand\copyrightnotice{%
\begin{tikzpicture}[remember picture,overlay]
\node[anchor=south,yshift=10pt] at (current page.south) {\fbox{\parbox{\dimexpr\textwidth-\fboxsep-\fboxrule\relax}{\copyrighttext}}};
\end{tikzpicture}%
}
\usepackage{cite}
\usepackage{amsmath,amssymb,amsfonts}
\usepackage{algorithmic}
\usepackage{graphicx}
\usepackage{textcomp}
\usepackage{xcolor}
\usepackage{dblfloatfix}
\ifCLASSOPTIONcompsoc
\usepackage[caption=false,font=normalsize,labelfon
t=sf,textfont=sf]{subfig}
\else
\usepackage[caption=false,font=footnotesize]{subfi
g}
\fi

\def\BibTeX{{\rm B\kern-.05em{\sc i\kern-.025em b}\kern-.08em
    T\kern-.1667em\lower.7ex\hbox{E}\kern-.125emX}}
\begin{document}

\title{ Toward industrial use of continual learning : new metrics proposal for class incremental learning\\
}

\author{\IEEEauthorblockN{Mohamed Abbas Konaté\IEEEauthorrefmark{1}\IEEEauthorrefmark{2}, Anne-Françoise Yao\IEEEauthorrefmark{2}, Thierry Chateau\IEEEauthorrefmark{3}, Pierre Bouges\IEEEauthorrefmark{1}}
\\
\IEEEauthorblockA{\IEEEauthorrefmark{1} Michelin\\
\{mohamed-abbas.konate, pierre.bouges\}@michelin.com}

\IEEEauthorblockA{\IEEEauthorrefmark{2} Université Clermont Auvergne, CNRS, Laboratoire de Mathématiques Blaise Pascal, 63000, France\\
anne.yao@uca.fr}
\IEEEauthorblockA{\IEEEauthorrefmark{3} Université Clermont Auvergne,Clermont Auvergne INP, CNRS, Institut Pascal,63000 Clermont–Ferrand, France \\
thierry.chateau@uca.fr}

}
%

\maketitle
\copyrightnotice
\begin{abstract}
In this paper, we investigate continual learning performance metrics used in class incremental learning strategies for continual learning (CL) using some high performing methods.
We investigate especially mean task accuracy. First, we show that it lacks of expressiveness through some simple experiments to capture performance. We show that monitoring average tasks performance is over optimistic and can lead to misleading conclusions for future real life industrial uses.
Then, we propose first a simple metric, Minimal Incremental Class Accuracy (MICA) which gives a fair and more useful evaluation of different continual learning methods.
Moreover, in order to provide a simple way to easily compare different methods performance in continual learning, we derive another single scalar metric  that take into account the learning performance variation as well as our newly introduced metric.
\end{abstract}


\section{Introduction}

Deep neural networks use has been booming during this last decade, performing tremendously in many domains, such as computer vision, natural language processing. Industrial applications have been multiplying since it offers fantastic performance gap sometimes \cite{he_deep_2015}.
In more traditional industries, especially manufacturing companies, huge transformations have been put in motion to make the companies use more data, and incidentally deep learning models in their process.

At the same time, economics behaviors of customers shift rapidly, with always more personalized items and goods. As habits changes rapidly, industrial models which powers now these products need to change quickly and always do more tasks \cite{deloitte_deloitte_nodate}.

An elegant response to this challenge would be to give the models the ability to evolve like humans, in particular learn and memorize in a continuous manner. Here is the essence of continual learning (also referred as incremental learning), a paradigm which aims at creating deep learning methods that can continuously learn new knowledge while retaining old knowledge.

This is a challenge as we know since \cite{mccloskey_catastrophic_1989} that deep learning models suffer from catastrophic forgetting. Catastrophic forgetting is the tendency for deep neural networks to forget previous knowledge when presented new ones.
This is because deep learning model trained with backpropagation are extremely plastic and tend to solely optimize toward the current goal.

\begin{figure}[!t]
\centering
\includegraphics[width=0.5\textwidth]{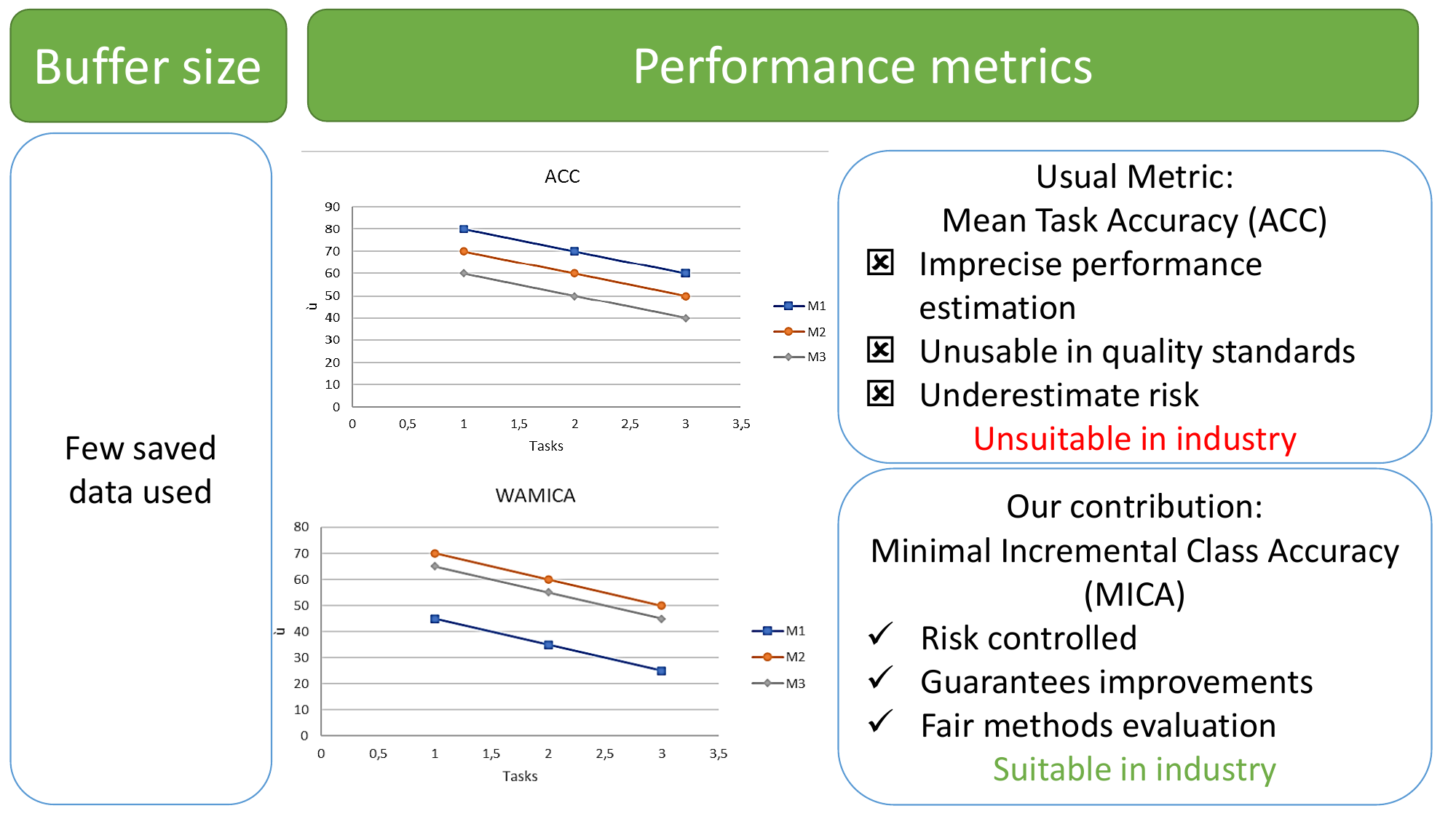}
\caption{We investigate how the Mean Task Accuracy impacts industrial equipements performance. We find that this metrics underestimate the risk of failure and can lead to wrong choices of methods. We correct this problem with our new metric, Minimal Incremental Class Accuracy which gives a lower bound of the performance. Our metric guarantee fairness in CIL methods comparison and is safe for a quality management system in industrial cases.}
\label{fig_cifar10_20}
\end{figure}

Continual learning topic is gaining more attention lately. Very promising works have already been done on the topic\cite{mundt_cleva-compass_2021}.
In this work, we focus on the classification problem as it is one of the most studied issue.
The research have been structured in 3 mains directions, Class Increment Learning (CIL), Task Incremental Learning (TIL) and Domain Incremental Learning (DIL)\cite{masana_class-incremental_2021}.
Although TIL can be an answer in very specific uses cases in industrial areas, it remains an unrealistic use case as we need an oracle to guide the model toward a limited set of outputs to give a correct answer.
True continual learning in classification is much more addressed in CIL and will be our focus.

A very important requirement for any continual learning system is the ability to guarantee a known and validated level of performance during the whole lifetime of the model. This makes the right choice of the metric used to estimate the forgetting of the model crucial.
Usually, in industrial use cases, models have to be validated by quality management systems. A well-established method to evaluate the failure of a system is to run the Failure Mode, Effects, and Criticality Analysis (FMECA). It is normalised in norms like MIL-STD 1629 or BS 5760-5 \cite{center_us_failure_1993}.
This method aims at finding  any potential failure, and estimating the criticity of the considered industrial system. These failures can come from any components in the system.
Criticity is then defined as the product of 3 factors : 
\begin{itemize}
\item the gravity of the failure mode
\item the frequency of apparition 
\item the ability to detect the failure during the manufacturing process
\end{itemize}
As a component of a industrial system, CIL Models will be subjected to this analysis. They have to provide a fair evaluation of their prediction performance. In particular, when used to replace quality operations, their ability to correctly estimate their own leak rate for the detection is critical.
This is mandatory to correctly estimate the failure risk of the system depending on the AI during the FMECA analysis, hence define correctly the acceptable risk taken by the company.

Through extensive bibliographic research, we noted some limits in the way CIL algorithms are evaluated. This can be misleading in assessment of their performance in real-lifes uses cases. Our contribution lies in this domain:

\begin{itemize}
\item First, we analyse current performance metrics used and show their lack of expressiveness to capture true performance, then propose a more restrictive and significant metric that can solve this problem.
\item Secondly, since we will use such metric systems for decisions making, we propose a new way to compute a single scalar performance value to easily compare differents methods for CIL methods performance assessment. 

\end{itemize}

In the rest of the paper, we will discuss the different families of strategies in CIL, then expose the choice of some strategies above others as references for this work. Next, we will analyse the results of training experiments in the light of classical metrics, to finally introduce our two metrics which correct the weaknesses of classically used metrics.
\section{Previous works}

\subsection{Continual learning methods}
A lot of work have been done to tackle class incremental learning in classification and the last years have an increasing trend in the research on the domain \cite{mundt_cleva-compass_2021}. Most of the discovered methods can be sorted in 3 mains directions, regularisation methods, dynamic architectures  and rehearsal methods. As studied in \cite{van_de_ven_three_2018}, regularisations methods like \cite{kirkpatrick_overcoming_2017} \cite{zenke_continual_2017} \cite{aljundi_memory_2018} tend to underperform in the class incremental learning setup when rehearsal methods seems to be more suited to this continual learning setup.
Rehearsal methods like \cite{rebuffi_icarl_2017}\cite{douillard_podnet_2020}\cite{zhao_maintaining_2019}\cite{de_lange_continual_2021-1} \cite{aljundi_online_2019} consist in keeping or generating past data to augment the new classes data during training.
Old data used in rehearsal methods can be saved from previous dataset. Many strategies have been proposed to choose and update this buffer \cite{rebuffi_icarl_2017}\cite{aljundi_online_2019}.
Dynamic architectures \cite{rusu_progressive_2016}\cite{yan__2021}\cite{yoon_lifelong_2018}\cite{sokar_spacenet_2021} aim at modifying the structure of the model to fight catastrophic learning. This is done usually in at least two stages. First, expanding (selectively or not) and training the neural network, then pruning such models.

\textbf{As it is the most interesting in industrial scenarios, we study the case where there is very limited access to previous data }. This case may happen for example when for legal reason the data cannot be stored. The data also may have save expiration date because of enterprise content management policies. Finally, a common case can be when provided by external supplier or partner a neural network already trained but no access to the data used for training.

\subsection{Continual learning metrics}
As introduced early, CIL in classification aims at learning differents classes appearing in a stream of data.
Let assume a class incremental learning setup composed of \textit{T} tasks.

A task is a set of \textit{C} classes that have to be learnt together. We note $C_{j}$, the number of classes in the task $t_{j}$ for $j \in \{1..T\}$.\\
To estimate model performance, we assume the existence of a test set containing all classes contained in all T tasks . This assumption implications will be discussed in \ref{discussionSection}.

Many metrics have been are currently used in CIL performances accessment. The works \cite{lopez-paz_gradient_2017}\cite{diaz-rodriguez_dont_2018}\cite{farquhar_towards_2019} \cite{kemker_measuring_2018} describes extensively these metrics and can be used to get a wide overview of used metrics. \cite{farquhar_towards_2019} in particular criticizes the design of current experiments to evaluate continual learning methods. In our work, we focus on performance accessment in the decision making process of choosing one method over another for CIL use cases.

\textbf{Mean Task Accuracy}: The most used one for performance accessment is the Mean Task Accuracy on all seen tasks. This metric is sometimes called Mean Incremental Accuracy in some works and refered to in short as accuracy (ACC). We will keep this abbreviation for the rest of the paper.

Given the test accuracy matrix R where $R_{ij}$ is the mean of classes accuracies for classes part of task $t_{j}$, after the end of the training on task $t_{i}$.
$R_{ij}$ is defined for $j \leq i$ always.
\begin{equation}
R=
\begin{bmatrix}
R_{11} & -  &  \dots & - \\
 R_{21} & R_{22} &  \dots & - \\
\vdots & \vdots & \ddots  & - \\
 R_{T1} &  R_{T2} & \dots &  R_{TT} \\
\end{bmatrix}, 
\label{confusion_matrix}
\end{equation}

Given $ r_{ijk}$, the accuracy of the class $c_{k}$, part of task $t_{j}$ and computed after task $t_{i}$, we have :  \\  
\begin{equation}
R_{ij}=\frac{1}{C_{j}}\sum_{k=1}^{C_{j}}r_{ijk}, k \in [[1..C_{j}]]
\end{equation}

This metric computes after training task $t_{i}$, the average of the test accuracies $R_{j}$ of each task $t_{j}\leq t_{i}$.
\begin{equation}
ACC_{i}=\frac{1}{i}\sum_{j=1}^{i}R_{ij}, i \in [[1..T]]
\end{equation}

Another commonly used metric is backward transfer (BWT) introduced in  \cite{lopez-paz_gradient_2017} and refined in \cite{diaz-rodriguez_dont_2018}. This metric aims at measuring the influence of learning a new task have on the previous tasks.
Formaly, after training task $t_{i}$,
\begin{equation}
BWT_{i}=\frac{1}{i-1}\sum_{j=1}^{i-1}(R_{ii}-R_{ij}), i \in [[1..T]]
\end{equation}

\subsection{Experiments description}
In order to analyse the expressiveness of Mean Task Accuray (ACC), we design an experiment defined in the following:\\
\begin{itemize}
\item \textbf{A1 - On CIFAR100,  10 tasks of 10 disjoints classes} \label{benchmarkDefcifar100}
\item \textbf{A2 - On CIFAR10, 5 tasks of 2 disjoints classes} \label{benchmarkDefcifar100}
\end{itemize}
This is a very common setup used in many works and it is used in the chosen works to compare themslves performance to other CIL works.\\
We choose 3 strategies for CIL, 2 of the recent and high ACC performing methods on CIL and Gdumb as a weak baseline. 
In detail, the chosen strategies are:
\begin{itemize}
\item \textbf{Gdumb}  \cite{prabhu_gdumb_2020}, a naive method that is surprinsingly very competitve with many approches in CIL. This method will act as our naive method to fight catastrophic learning.
\item \textbf{Dynamic Expandable Representation (DER)} \cite{yan__2021}, one hybrid method which mixes both architectural and rehearsal ideas. This method concatenate many networks in one to mitigate forgeting. To do so, it freezes the old model weights  while learning a representation of new task. Then, 
to bias in classification layer; it retrains a balanced subset of all classes at the end of the training. A optional step of pruning can be implemented to decrease the memory footprint of the model with minimal drop in \textit{ACC}.
In the original work, it achieves an impressive 75.36 \% $\pm$ 0.36 ACC on the \textbf{A1} setup for a constant buffer of 2000 samples.
We choose this method because it seems to performs very well on catastastrophic learning. As a matter of fact, it is ranked first on \cite{noauthor_papers_nodate} for the A1 experiment.
\item\textbf{Weight Align (WA)} \cite{zhao_maintaining_2019}, a pure rehearsal method that introduces a smart trick to mitigate bias in classification layer to boost knowledge distillation-based methods for CIL. It achieves 69.46 \% $\pm$ 0.29 ACC on the \textbf{A1} setup for 2000 saved samples.
This method by it simplicity and high performance ACC can be a very good compromise in a future industrial use.
\end{itemize}

Since we aim at working with no to very rare data, we define buffer management as a fixed buffer size per classe of either 2 or 20 samples.

The buffer have been managed according to the Nearest Mean Exemplars technique defined in \cite{rebuffi_icarl_2017}. 
We trained all methods with the modified version of Resnet18 for CIFAR introduced in \cite{rebuffi_icarl_2017} with 3 runs. The class order are the one followed by \cite{yan__2021} for CIFAR100 and randomly generated (but kept fixed for all methods) for CIFAR10.
\subsection{Experiments analysis }

\begin{figure*}[!t]
\centering
\subfloat[Cifar100 : 10 tasks of 10 classes ]{\includegraphics[width=0.7\textwidth]{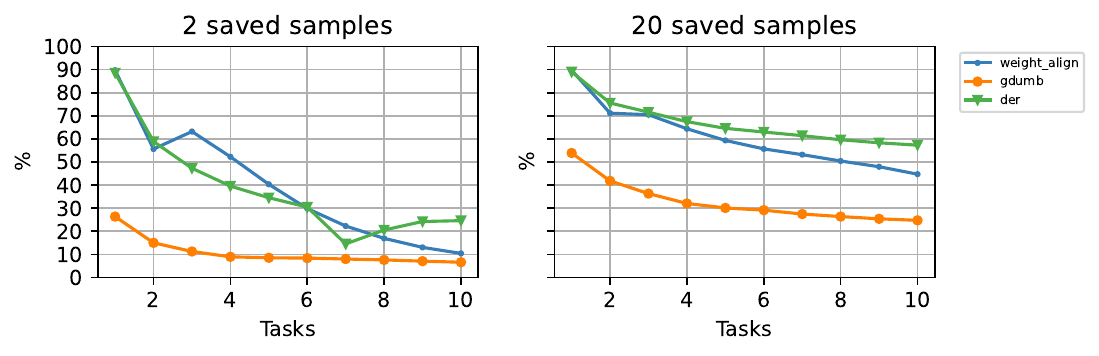}
\label{fig_first_case}}
\vfil
\subfloat[Cifar 10 : 5 tasks of 2 classes]{\includegraphics[width=0.7\textwidth]{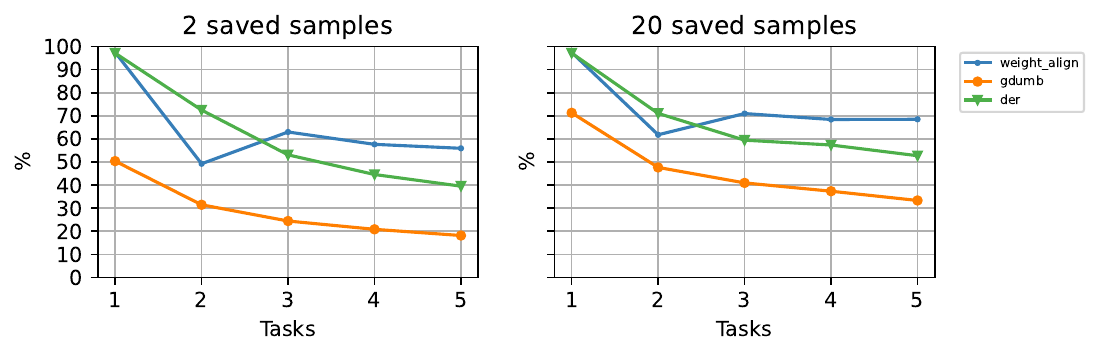}
\label{fig_second_case}}
\caption{Mean Task Accuracy (ACC)}
\label{acc_ref}
\end{figure*}

\begin{table}[!t]
\renewcommand{\arraystretch}{1.3}
\caption{CIFAR100 : Mean Task Accuracy (ACC) \%}
\label{cifar100_table_acc}
\centering
\begin{tabular}{|c||c||c|}
\hline
\bfseries \#Saved samples per class  &\bfseries 2 &\bfseries 20 \\
\hline\hline
Gdumb &10.78 &32.73 \\

DER & 38.33 & \textbf{66.77}\\

WA  &\textbf{39.37} & 60.70\\
\hline
\end{tabular}
\end{table}

\begin{table}[!t]
\renewcommand{\arraystretch}{1.3}
\caption{CIFAR10 : Mean Task Accuracy (ACC) \%}
\label{cifar10_table_acc}
\centering
\begin{tabular}{|c||c||c|}
\hline
\bfseries \#Saved samples per class  &\bfseries 2 &\bfseries 20 \\
\hline\hline
Gdumb &29.07 & 46.12 \\

DER &61.37 &67.34\\

WA  & \textbf{64.67} &\textbf{ 73.42}\\
\hline
\end{tabular}
\end{table}

As we can see in table \ref{cifar10_table_acc}, table \ref{cifar100_table_acc}, showing the ACC of all tested methods on incremental CIFAR100 and 10 datasets,  and Fig. \ref{acc_ref}, highlighting the test curve per tasks, in the case of CIFAR100, one would probably pick DER as the overall best performing method, and for CIFAR10, Weight Align method. The worst strategy would be Gdumb.  But here, we look at the average incremental performance of tasks.

A fair question to ask oneself would be : \\ 
\textit{ Is this mean accuracy representative of all classes accuracy performance learnt in the differents tasks seen yet ? }\\
This response to this question is critical to jugde correcty these results. 
Instead of the ACC, we are interested in the distribution of classes accuracies.

\begin{figure*}[!t]
\centering
\subfloat[Cifar 100  : 20 samples]{
\includegraphics[width=0.5\textwidth]{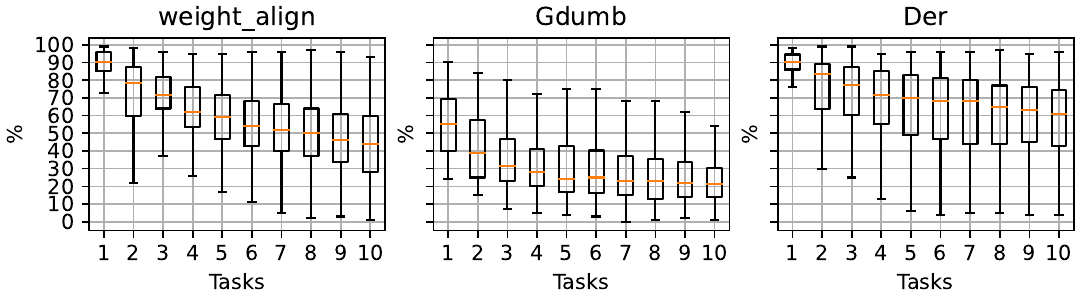}
\label{fig_cifar10_202_mean}}
\subfloat[Cifar 100 :2 samples]{
\includegraphics[width=0.5\textwidth]{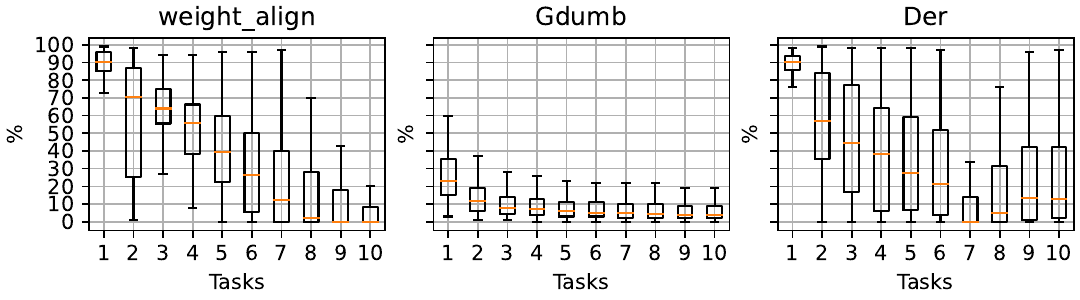}
\label{fig_cifar100_202_mean}}

\subfloat[Cifar 10  :20 samples]{
\includegraphics[width=0.5\textwidth]{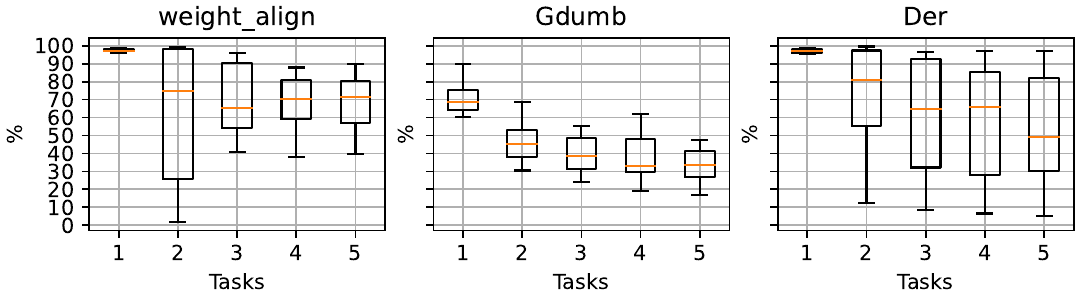}
\label{fig_cifar10_202_mean}}
\subfloat[Cifar 10 :2 samples]{
\includegraphics[width=0.5\textwidth]{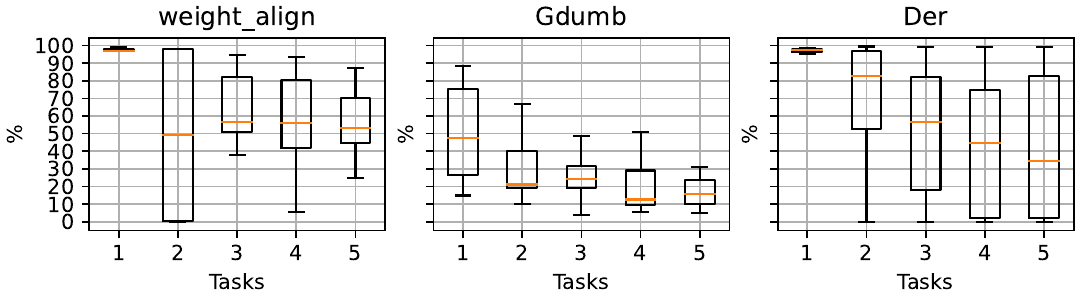}
\label{fig_cifar100_202_mean}}
\caption{Distribution of classes accuracies during training}
\label{distrib_boxplot_20}

\end{figure*}

We produced the boxplot of these training sessions to get a less compressed view of the training process. The result is given by Fig. \ref{distrib_boxplot_20}.
We can see very large variations of the distributions of classes accuracies for all methods, especially the 2 high performing ones. This clearly indicates that the accuracies between all classes varies a lot. These variations get even worse when we reduce the number of saved samples.\\
These large variations are quite problematic for a future use of such methods to tackle CIL problems.
For example, in a quality management system based on ACC, a method which exhibit such large variations between defects detection could lead to a surge of reclamations, while acceptable risk leakage threshold might no be attained according to the metric.

Not only ACC fails to give us a good evaluation of the performance taking the average performance, but it does not inform us on the distributions of the other classes around it.
This ascertainment lead us to formulate another metric much more suited for the continual learning problem.

\section{Worst case metric dimensioning}
\subsection{A more informative metric: Minimum Incremental Class Accuracy: MICA}
As in mechanical engineering, where worst case dimensioning is a conception rule, we advocate to have a clear idea of the performance of such methods, we should always be pessimistic instead of being too optimistic on performances can lead to dramatic mistakes in real- life uses.
The performance of such model after each task should be viewed through the worst performing class, instead of the average of all classes performances.\\ We propose then a new simple metric to use : the \textbf{Minimum Incremental Class Accuracy}.

This metric compute after training task $t_{i}$, the minimum of the classes accuracies $r_{ijk}$ of each of the seen classes $c_{k} \in \cup\{ t_{j}, j \leq i \}$, from task $t_{1}$ to current  task $ t_{i}$.

\begin{equation}
MICA_{i}=min(r_{ijk})
\end{equation}

\begin{figure*}[!t]
\centering
\subfloat[Cifar100 : 10 tasks of 10 classes ]{\includegraphics[width=0.6\textwidth]{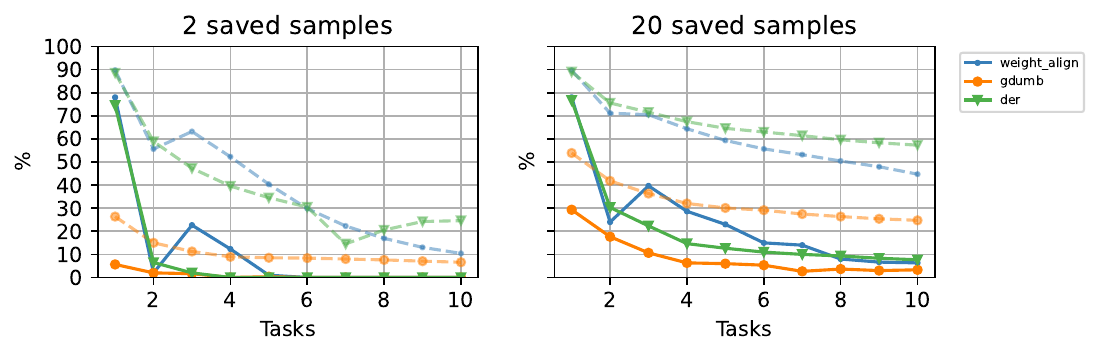}
\label{fig_first_case}}
\vfil
\subfloat[Cifar 10 : 5 tasks of 2 classes]{\includegraphics[width=0.6\textwidth]{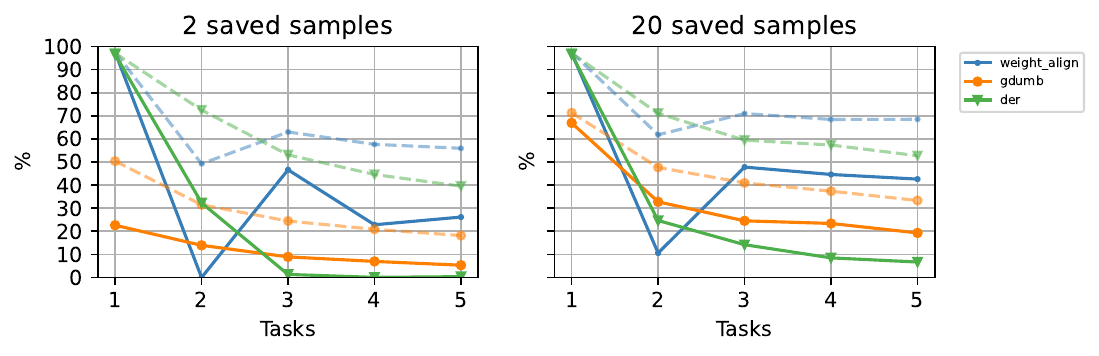}
\label{fig_second_case}}
\caption{Minimum Incremental Class Accuracy ( MICA) (in transparent dashed, ACC)}
\label{mica}
\end{figure*}

This is a more restrictive metric which is directly linked to the true expected performance of a method. Increasing this metric value ensures that all other classes learnt (new and old) are at least better than this lower bound value.
One can decide an usability threshold depending on the application needs.
Using this metric, we compare  in Fig. \ref{mica} with more confidence these methods and garantee true improvements over state of the art methods.
In the case of CIFAR100, it appears that sophisticated method like DER or Weight Align perform similar to Gdumb when the number of tasks increases or the number of saved samples goes down. Especially, DER do not seems to be good at fighting catastrophic learning as it perform worse than Gdumb when put in very rare data use cases.

Moreover, another key information is how well the old classes behave compared to new introduced classes. For the same reasons exposed before, backward transfer (BWT) as defined in \cite{lopez-paz_gradient_2017} and \cite{diaz-rodriguez_dont_2018} could lead to mistakes in performance accessment. 
This information, called \textit{remembering} in \cite{diaz-rodriguez_dont_2018} should be exposed in a clear way so that one can understand wether one method is perfoming better than another solely on castastrophic learning.
We propose to used again MICA, but only restricted to the old classes to access the performance on catastrophic learning.\\

\subsection{Going Futher : One value to rule them all}
Although not adapted to the incremental setup, the mean incremental accuracy (ACC) have the advantage of summarizing in one scalar all the training process. It offers a convenient way to compare differents methods without requiring the whole test curve like in Fig \ref{mica}. 
We wish to provide such convenient scalar value to easily compare the methods. We could compute the mean of our metric MICA, but we would fall in the same caveat of ACC, i.e., not taking in account the variation of performance for classes between the first learned task and the last one.
A better metric should be able to integrate the variation of MICA between the first task and the last task considered then. 

We propose a \textbf{weighted average of MICA} (WAMICA). Let define $MICA_{min}$ and $MICA_{max}$ respectively the minimum and the maximum of  MICA of all incremental T tasks.
This metric $WAMICA$ is composed of 2 terms. First, the mean of the MICA of all incremental T tasks, times a weight $w_{T}$  define as :
\begin{equation}  \label{ wmact}
w_{T} =  1 - (MICA_{max}- MICA_{min})
\end{equation}

Then, 
\begin{equation}  \label{ mmact}
WAMICA= (\frac{w_{T}}{T}\sum_{j=1}^{T}MICA_{j}) 
\end{equation}

\begin{figure*}[!t]
\centering
\includegraphics[width=0.4\textwidth]{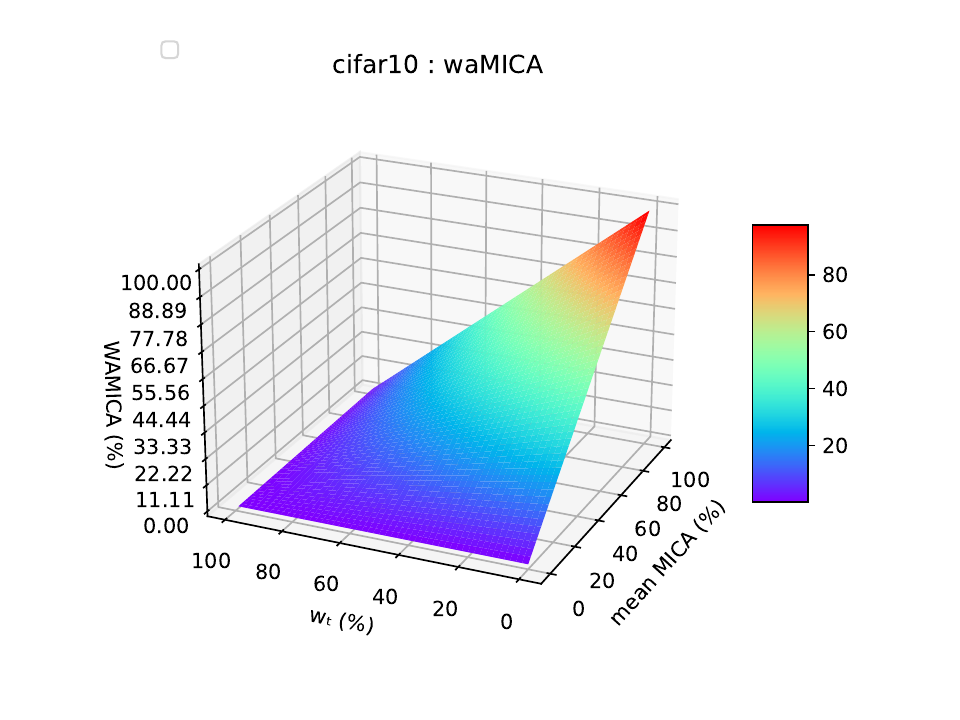}
\caption{Weighted Average Minimal Incremental Class Accuracy (best viewed in color)}.
\label{wamica_ref}
\end{figure*}

As we can see on Fig. \ref{wamica_ref}, this metric take into account the variations of the tasks metrics through $w_{T}$. If we have a very dispersed tasks MICA, it will penalize the overall mean MICA performance. 
This gives us a better view of how a CIL method is performing on the whole spectrum of encountered tasks.

It is easy now to show using tables \ref{cifar100_table_wamica} and  \ref{cifar10_table_wamica} that all the chosen methods have actually poor and similar performance as they all lie in the bottom of the metric surface, despite the good performance achieved on ACC.\\
Actually, we can see that the most effective metric for the benchmark A1 and A2 is Gdumb. This result clearly backup our initial hypothesis that ACC is not very well suited to describe the performance of such systems.

\begin{table}[!t]
\renewcommand{\arraystretch}{1.3}
\caption{CIFAR100 : weighted average Minimal Incremental Class Accuracy (WAMICA) \%}
\label{cifar100_table_wamica}
\centering
\begin{tabular}{|c||c|c||c|c|}
\hline
\bfseries \# Saved samples per class &\bfseries 2 &\bfseries 2 &\bfseries 20 &\bfseries 20 \\
\hline\hline
\bfseries Metrics &\bfseries ACC  &\bfseries WAMICA &\bfseries ACC &\bfseries WAMICA \\
\hline
Gdumb & 10.78 & 0.91 & 32.73 & 6.45 \\
DER & 38.33& 2.1 & \bfseries 66.77 & 6.29\\
WA  & \bfseries 39.37 & \bfseries 2.5 & 60.70 & \bfseries  6.89\\
\hline
\end{tabular}
\end{table}

\begin{table}[!t]
\renewcommand{\arraystretch}{1.3}
\caption{CIFAR10 : Weighted Average Minimal Incremental Class Accuracy (WAMICA) \%}
\label{cifar10_table_wamica}
\centering
\begin{tabular}{|c||c|c||c|c|}
\hline
\bfseries \# Saved samples per class &\bfseries 2 &\bfseries 2 &\bfseries 20 &\bfseries 20 \\
\hline
\bfseries Metrics &\bfseries ACC  &\bfseries WAMICA &\bfseries ACC &\bfseries WAMICA \\
\hline\hline
Gdumb &29.07 & \bfseries 9.56 & 46.12& \bfseries 17.4 \\

DER & 61.37& 0.9 & 67.34 & 3.04\\

WA  &\bfseries 64.67&  0.9 & \bfseries73.42 & 6.5\\
\hline
\end{tabular}
\end{table}

\section{Discussions} \label{discussionSection}
A key assumption in this work and many others is the fact that there exists a test set which can be used to access the performance of our method on our dataset. 
This assumption seems quite strong because we have no such guarantee in practice. Continual learning will deliver full help in rare to no data problems. Hence, if there is no old data saved, or a very little amount of them, it may be impossible or not desirable to keep on hand data that could be useful during training for testing purposes. 

Secondly, when evaluating a CIL method, we would like to be able to separate the effect of the method internal working and the effect of the dataset and classes order of appearance on the performance. When working on datasets like CIFAR10/1000 or ImageNet, we might shuffle the dataset to be agnostic to these variations.
But in real-lifes scenarios, we don't have such luxury to pick the order of the classes we have to continuously learn. Hence continual learning methods are very dataset dependent in practice. It is probable that some poor performing methods on some datasets might perform better on others. This highlights the importance of having a simple and reliable way like WAMICA to quickly compare methods.

Finally, while continual learning topic is booming, it's still not a mature domain. A fair criticism to our work could be that current methods are far from being applicable in real life use, hence the metric we propose could be too restrictive for the need of the domain right now. Although we share the first part of point of view, we believe that it will be with right metrics that true improvements can be achieved and monitored. Our metric MICA helps to answer that challenge.

\section{Conclusion and and future works} \label{conclusion}
In this paper, we demonstrated that currently used metric for performance acessement Mean Incremental Accuracy (ACC)  in Class Incremental Learning are misleading on the actual performance of methods. We run an extensive experiment on CIFAR10 and 100, with 3 recent high performing CIL with respect to ACC methods (DER,Gdumb and Weight Align) which show significant variations of performance in classes during learning. We then propose a new metric, Mean Incremental Class Accuracy (MICA) which, act as a lower bound guaranties that the model performs always at least better than the metric level. Based on this metric MICA, we propose another single metric scalar, WAMICA, that sumarize the evolution and the variation of model performance during the training process and can be used to safely compare differents CIL methods.

We purposely didnt' develop methods related to others interesting properties like memory and energy footprint, training and inference time and other non-accuracy related metrics used in CIL. It is clear that these metrics have to be taken in account when designing or choosing a method for continual learning.
As exposed in \ref{discussionSection} also, it will be vital to be able to estimate catastrophic learning during training without any test sets help.
These challenges will be  key research directions for our next works.

\section*{Acknowledgments}
This work made in the FACTOLAB (commun laboratory CNRS, UCA, Michelin) and International Research Center " Innovation Transportation and Production Systems" of the I-SITE CAP 20-25 frameworks was supported by Michelin Tyres Manufacturer.

\bibliographystyle{plain}
\bibliography{ContinualLearningKMA,ContinualLearningPapers}

\begin{thebibliography}{10}

\bibitem{noauthor_papers_nodate}
Papers with {Code} - {CIFAR100}-{B0}(10steps of 10 classes) {Benchmark} ({Incremental} {Learning}).

\bibitem{aljundi_memory_2018}
Rahaf Aljundi, Francesca Babiloni, Mohamed Elhoseiny, Marcus Rohrbach, and Tinne Tuytelaars.
\newblock Memory {Aware} {Synapses}: {Learning} what (not) to forget.
\newblock In {\em The {European} {Conference} on {Computer} {Vision} ({ECCV})}, September 2018.

\bibitem{aljundi_online_2019}
Rahaf Aljundi, Lucas Caccia, Eugene Belilovsky, Massimo Caccia, Min Lin, Laurent Charlin, and Tinne Tuytelaars.
\newblock Online {Continual} {Learning} with {Maximally} {Interfered} {Retrieval}.
\newblock {\em arXiv:1908.04742 [cs, stat]}, October 2019.
\newblock arXiv: 1908.04742.

\bibitem{center_us_failure_1993}
Reliability~Analysis Center~(U.S.), M.J. Rossi, R.J. Borgovini, S.~Pemberton, RELIABILITY ANALYSIS CENTER GRIFFISS~AFB NY, and N.Y.) Rome Laboratory (Griffiss Air Force~Base.
\newblock {\em Failure {Mode}, {Effects} and {Criticality} {Analysis} ({FMECA})}.
\newblock Concurrent engineering series. The Center, 1993.

\bibitem{de_lange_continual_2021-1}
Matthias De~Lange and Tinne Tuytelaars.
\newblock Continual {Prototype} {Evolution}: {Learning} {Online} from {Non}-{Stationary} {Data} {Streams}.
\newblock pages 8250--8259, 2021.

\bibitem{deloitte_deloitte_nodate}
Deloitte.
\newblock The {Deloitte} {Consumer} {Review} {Made}-to-order: {The} rise of mass personalisation.

\bibitem{douillard_podnet_2020}
Arthur Douillard, Matthieu Cord, Charles Ollion, Thomas Robert, and Eduardo Valle.
\newblock {PODNet}: {Pooled} {Outputs} {Distillation} for {Small}-{Tasks} {Incremental} {Learning}.
\newblock {\em European Conference on Computer Vision (ECCV)}, 2020.

\bibitem{diaz-rodriguez_dont_2018}
Natalia Díaz-Rodríguez, Vincenzo Lomonaco, David Filliat, and Davide Maltoni.
\newblock Don't forget, there is more than forgetting: new metrics for {Continual} {Learning}.
\newblock {\em arXiv}, October 2018.

\bibitem{farquhar_towards_2019}
Sebastian Farquhar and Yarin Gal.
\newblock Towards {Robust} {Evaluations} of {Continual} {Learning}.
\newblock In {\em Privacy in {Machine} {Learning} and {Artificial} {Intelligence} workshop, {ICML}}, June 2019.

\bibitem{he_deep_2015}
Kaiming He, Xiangyu Zhang, Shaoqing Ren, and Jian Sun.
\newblock Deep {Residual} {Learning} for {Image} {Recognition}, December 2015.
\newblock arXiv:1512.03385 [cs].

\bibitem{kemker_measuring_2018}
Ronald Kemker, Marc McClure, Angelina Abitino, Tyler~L Hayes, and Christopher Kanan.
\newblock Measuring {Catastrophic} {Forgetting} in {Neural} {Networks}.
\newblock In {\em Thirty-{Second} {AAAI} {Conference} on {Artificial} {Intelligence}}, April 2018.

\bibitem{kirkpatrick_overcoming_2017}
J.~Kirkpatrick, Razvan Pascanu, Neil~C. Rabinowitz, J.~Veness, Guillaume Desjardins, Andrei~A. Rusu, K.~Milan, John Quan, Tiago Ramalho, Agnieszka Grabska-Barwinska, D.~Hassabis, C.~Clopath, D.~Kumaran, and R.~Hadsell.
\newblock Overcoming catastrophic forgetting in neural networks.
\newblock {\em Proceedings of the National Academy of Sciences}, 114:3521 -- 3526, 2017.

\bibitem{lopez-paz_gradient_2017}
David Lopez-Paz and Marc'Aurelio Ranzato.
\newblock Gradient {Episodic} {Memory} for {Continual} {Learning}.
\newblock 2017.

\bibitem{masana_class-incremental_2021}
Marc Masana, Xialei Liu, Bartlomiej Twardowski, Mikel Menta, Andrew~D. Bagdanov, and Joost van~de Weijer.
\newblock Class-incremental learning: survey and performance evaluation on image classification.
\newblock {\em arXiv:2010.15277 [cs]}, May 2021.
\newblock arXiv: 2010.15277.

\bibitem{mccloskey_catastrophic_1989}
M.~McCloskey and N.~Cohen.
\newblock Catastrophic {Interference} in {Connectionist} {Networks}: {The} {Sequential} {Learning} {Problem}.
\newblock {\em Psychology of Learning and Motivation}, 24:109--165, 1989.

\bibitem{mundt_cleva-compass_2021}
Martin Mundt, Steven Lang, Quentin Delfosse, and Kristian Kersting.
\newblock {CLEVA}-{Compass}: {A} {Continual} {Learning} {Evaluation} {Assessment} {Compass} to {Promote} {Research} {Transparency} and {Comparability}.
\newblock September 2021.

\bibitem{prabhu_gdumb_2020}
Ameya Prabhu, Philip H.~S. Torr, and Puneet~K. Dokania.
\newblock {GDumb}: {A} {Simple} {Approach} that {Questions} {Our} {Progress} in {Continual} {Learning}.
\newblock In Andrea Vedaldi, Horst Bischof, Thomas Brox, and Jan-Michael Frahm, editors, {\em Computer {Vision} – {ECCV} 2020}, Lecture {Notes} in {Computer} {Science}, pages 524--540, Cham, 2020. Springer International Publishing.

\bibitem{rebuffi_icarl_2017}
Sylvestre-Alvise Rebuffi, Alexander Kolesnikov, G.~Sperl, and Christoph~H. Lampert.
\newblock {iCaRL}: {Incremental} {Classifier} and {Representation} {Learning}.
\newblock {\em 2017 IEEE Conference on Computer Vision and Pattern Recognition (CVPR)}, pages 5533--5542, 2017.

\bibitem{rusu_progressive_2016}
Andrei~A Rusu, Neil~C Rabinowitz, Guillaume Desjardins, Hubert Soyer, James Kirkpatrick, Koray Kavukcuoglu, Razvan Pascanu, and Raia Hadsell.
\newblock Progressive {Neural} {Networks}.
\newblock {\em arXiv}, June 2016.

\bibitem{sokar_spacenet_2021}
Ghada Sokar, Decebal~Constantin Mocanu, and Mykola Pechenizkiy.
\newblock {SpaceNet}: {Make} {Free} {Space} for {Continual} {Learning}.
\newblock {\em Neurocomputing}, 439:1--11, June 2021.

\bibitem{van_de_ven_three_2018}
Gido~M van~de Ven and Andreas~S Tolias.
\newblock Three scenarios for continual learning.
\newblock In {\em Continual {Learning} {Workshop} {NeurIPS}}, 2018.

\bibitem{yan__2021}
Shipeng Yan, Jiangwei Xie, and Xuming He.
\newblock {DER}: {Dynamically} {Expandable} {Representation} for {Class} {Incremental} {Learning}, March 2021.
\newblock arXiv:2103.16788 [cs].

\bibitem{yoon_lifelong_2018}
Jaehong Yoon, Eunho Yang, Jeongtae Lee, and Sung~Ju Hwang.
\newblock Lifelong {Learning} with {Dynamically} {Expandable} {Networks}.
\newblock {\em arXiv:1708.01547 [cs]}, June 2018.
\newblock arXiv: 1708.01547.

\bibitem{zenke_continual_2017}
Friedemann Zenke, Ben Poole, and Surya Ganguli.
\newblock Continual {Learning} {Through} {Synaptic} {Intelligence}.
\newblock In {\em International {Conference} on {Machine} {Learning}}, pages 3987--3995, July 2017.

\bibitem{zhao_maintaining_2019}
Bowen Zhao, Xi~Xiao, Guojun Gan, Bin Zhang, and Shutao Xia.
\newblock Maintaining {Discrimination} and {Fairness} in {Class} {Incremental} {Learning}, November 2019.
\newblock arXiv:1911.07053 [cs].

\end{thebibliography}

%

\end{document}